# ProFlow: Learning to Predict Optical Flow

Daniel Maurer
maurer@vis.uni-stuttgart.de
Andrés Bruhn
bruhn@vis.uni-stuttgart.de

Institute for Visualization and
Interactive Systems
University of Stuttgart, Germany

**Abstract**

Temporal coherence is a valuable source of information in the context of optical flow estimation. However, finding a suitable motion model to leverage this information is a non-trivial task. In this paper we propose an unsupervised online learning approach based on a convolutional neural network (CNN) that estimates such a motion model individually for each frame. By relating forward and backward motion these learned models not only allow to infer valuable motion information based on the backward flow, they also help to improve the performance at occlusions, where a reliable prediction is particularly useful. Moreover, our learned models are spatially variant and hence allow to estimate non-rigid motion per construction. This, in turns, allows to overcome the major limitation of recent rigidity-based approaches that seek to improve the estimation by incorporating additional stereo/SfM constraints. Experiments demonstrate the usefulness of our new approach. They not only show a consistent improvement of up to 27% for all major benchmarks (KITTI 2012, KITTI 2015, MPI Sintel) compared to a baseline without prediction, they also show top results for the MPI Sintel benchmark – the one of the three benchmarks that contains the largest amount of non-rigid motion.

## 1 Introduction

Estimating the apparent motion from a given image sequence is one of the fundamental problems in computer vision. Typically, one is thereby interested in computing the inter-frame displacement field between consecutive frames, the so-called optical flow. In order to solve this task, many methods rely solely on two frames; see e.g. the recent methods in [13, 14, 19, 31, 35]. While they allow to obtain good results in most cases, they do not allow an actual reasoning in the context of occlusions. Leveraging information from additional frames could help to overcome their limitation. This, however, requires to formulate explicit motion models that relate the sought displacement vector field to motion estimates from the past.

While simple models based on a temporally constant flow can be a valid choice in case of sufficiently small motion [16], more complex models are required in general scenarios with fast and non-rigidly moving objects. Unfortunately, as observed in [10, 32], finding such models is a highly non-trivial task. Thus, recent multi-frame approaches resort to the scenario of mostly rigid scenes in order to still be able to use temporal information [34]. Assuming a moving camera and multiple independently moving objects, they exploit temporal information only in rigid parts of the scene – by solving a multi-frame stereo/SfM problem [28] there. Although the latter strategy can be very beneficial w.r.t. occlusions, it typically requires a sufficient amount of ego-motion. Moreover, the benefit of temporal information is limited to the rigid parts of the scene. Hence, to exploit the full potential of such information, it would be desirable to come up with a strategy that (i) does not rely on a moving camera and that (ii) allows to leverage temporal information also in non-rigid parts of the scene.

**Contributions.** In this paper, we tackle both problems. Instead of assuming a moving camera that comes with rigidity constraints, we propose a novel optical flow method that learns suitable motion models based on a convolutional neural network (CNN). In this context, our contributions are fourfold: (i) In contrast to other approaches that train a network before the estimation, our approach learns the models *online*, i.e. during the estimation. (ii) Moreover, instead of relying on potentially unsuitable data sets with ground truth, our models are trained using initial flow estimates of the *actual sequence*. Such an unsupervised training offers the advantage that appropriate models can be learned for each sequence. (iii) Thirdly, our approach not only learns one model per sequence but one model for *each frame* of every sequence. Evidently, this results in a high degree of adaptability when it comes to a change of the scene content. (iv) Finally, the learned models are *spatially variant*, i.e. location dependent. This in turn addresses the problem of independently moving objects. Having learned such dedicated motion models eventually enables us to predict the forward flow from the backward flow. Thus it becomes possible to improve the estimation at locations where the forward flow is not available, e.g. in occluded regions. Experiments make the benefits of our novel method explicit. They not only show consistent improvements compared to a baseline approach without prediction but also very good results for all major benchmarks in general.

## 1.1 Related Work

In the following we discuss related work in the field of optical flow estimation. Thereby we first focus on multi-frame methods and then comment on learning-based approaches.

**Multi-Frame Approaches.** In order to improve the quality and the robustness of the estimation, multi-frame approaches typically rely on some kind of motion model that describes how objects/pixels are expected to move over time. In this context, recent approaches go far beyond a simple constant velocity model [16, 17, 33] by using constraints based on constant acceleration [5, 32], parametrized trajectories [10, 27] or a moving camera [34]. Moreover, to avoid a significant deterioration of the results in case the model turns out to be inappropriate, they typically allow deviations from the model either by formulating it as a soft constraint [5, 10, 32] or by restricting the estimation to locations where the assumed model is most likely to hold [32, 34]. Compared to most of the aforementioned methods, our methods differs in two ways: On the one hand, our approach does not use hand-crafted or geometric/rigid motion models but *learns* spatially varying mappings from the backward to the forward flow. On the other hand, our approach uses the learned motion models as a *hard constraint*, i.e. without any filtering and at all location where the backward flow provides additional information, e.g. at occlusions. The only two methods that are close in spirit to our method are the approaches [27] and [10] that learn temporal basis functions for long term trajectories via PCA from pre-computed tracks and flow fields, respectively. However, in contrast to these approaches that focus on a robust long term motion representation to perform dense tracking and non-rigid video registration, respectively, our method aims at the classical *short-term optical flow setting* and, hence, only our approach is able to provide *state-of-the-art results* for standard optical flow benchmarks.

**Learning Approaches.** Regarding learning approaches for optical flow estimation, one can basically distinguish two types of methods: pure learning-based methods and partially learning-based methods. Pure learning-based methods aim at learning an end-to-end relation between the input images and the corresponding flow field, typically via one or multiple stacked CNNs [7, 15, 24, 31]. While the overall learning process is quite time-consuming and typically requires a large amount of training data, the learned models allow to compute high quality flow estimates in real-time [15, 31]. Recently, also unsupervised learning approaches have been considered to tackle the lack of realistic training data; see e.g. [2, 20, 25, 36, 37]. They either replace the ground truth by a proxy ground truth computed with recent optical flow methods [37] or they propose a loss function that does not depend on the ground truth, i.e. by using an image-based registration error [2, 20, 25, 36, 37] or some kind of smoothness constraint [20, 25, 36]. Partially learning-based approaches on the other hand, are hybrid methods: They seek to combine the advantages of two worlds. While relying on a transparent global energy minimization framework, they make use of machine learning techniques to replace some difficult task during the modeling or the estimation. These tasks includes descriptor learning [4, 8, 30, 35], instance level segmentation [3], rigidity estimation [34], and semantic scene segmentation [29]. Although our approach is partially-learning based, since it embeds a CNN into a traditional optical flow pipeline [26], it is completely different from all aforementioned learning-based approaches. Not only that the learning step solves a *different problem*, i.e. it predicts a forward flow from a backward flow, also the training itself is completely different. It uses an unsupervised *online* approach that relies on initial flow estimates to train the network *individually* for each frame of the sequence at runtime instead of training the network once for an entire task based on previously collected set of training data. In this respect, our approach is also intrinsically different from the unsupervised methods listed above.

## 2 Our Approach

Let us start by giving a brief overview over the proposed method; see Fig. 1. Please note that, in contrast to classical two frame approaches, our method considers image triplets, i.e. the frames at times $t-1$, $t$, and $t+1$. This allows to compute the optical flow from the reference frame $t$ not only to the subsequent frame $t+1$ (forward flow) but also to the previous frame $t-1$ (backward flow). After we have estimated both flows fields with a conventional optical flow approach, we perform outlier filtering via a bi-directional consistency check. While this requires the additional computation of flow fields using the reversed frame order, it allows us to identify possibly occluded image regions. Based on locations where both the forward and the backward flow are available after filtering, we then learn a model that allows to predict the forward flow from the backward flow. To this end, we train a CNN such that it performs a regression from small backward flow patches to forward flow vectors. Using the trained network, we then predict a new forward flow field from the filtered backward flow. This provides additional information at those locations where only the backward flow is given, e.g. at occlusions. Finally, the new and the initial forward flow field are combined such that predictions are used if no initial forward flow is available. As a last step, we inpaint the combined flow field to obtain dense results and refine it to improve its accuracy.

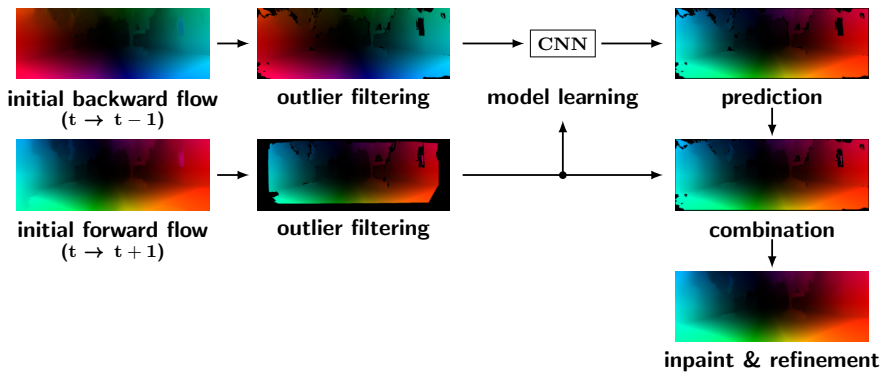

Figure 1: Schematic overview over our optical flow approach.

## 2.1　Initial Flow Estimation / Baseline

In a first step, we compute the initial forward and backward flow field, i.e. the flow fields from frame $t$ to frame $t+1$ and from frame $t$ to frame $t-1$, respectively. To this end, we consider a baseline approach that follows the four steps of the large displacement optical flow pipeline by Revaud *et al.* [26]: matching, outlier filtering, inpainting and variational refinement. However, instead of using the original components, we make use of recent progress in the field. For the matching we employ the coarse-to-fine PatchMatch approach (CPM) of Hu *et al.* [12], for the inpainting of the matches we use the robust interpolation technique (RIC) of Hu *et al.* [13] and for the final refinement we apply the order-adaptive illumination-aware refinement (OIR) of Maurer *et al.* [19]. Only the outlier filtering applied to the initial matches in terms of a bi-directional consistency check remains unchanged. Please note that this check requires to compute matches in the reverse direction as well, i.e. from frame $t+1$ to frame $t$ and from frame $t-1$ to frame $t$, respectively.

## 2.2　Outlier Filtering

After we have computed the initial forward and backward flow field with our baseline approach, we apply another outlier filtering step. Analogously to the bi-directional consistency check that is part of our baseline, this requires to compute flow fields in the reverse direction, i.e. from frame $t+1$ to frame $t$ and from frame $t-1$ to frame $t$, respectively. Once again, our baseline is used for this task. Finally, only those flow vectors are considered valid in the forward and backward flow field which are consistent with the corresponding vectors in the reverse direction. This allows to eliminate many outliers, in particular in occluded regions.

## 2.3　Learning a Motion Model

Having the filtered forward and backward flow field at hand, let us now discuss how the underlying motion model is learned. The goal of this step is to derive the relation between the backward flow and the forward flow which enables us to use the backward flow for predicting the forward flow at locations where the forward flow is not available. Since motion patterns typically vary across different scenes and frames, we do not use a network that has been trained in advance on a huge data base with ground truth [4, 15, 31, 35], but we apply an *unsupervised* learning approach that trains a CNN *during* the optical flow estimation – and that *individually* for each frame of the sequence. As shown in [9] in the context of predicting surface normals for multi-view stereo, such unsupervised learning techniques can be highly beneficial for densifying initially sparse results.

**Training Data Extraction.** The training data required for the learning process is extracted from the initially computed flow fields after outlier filtering. Thereby, all locations where both the forward and the backward flow surpassed the outlier filtering serve as potential training samples. These potential samples are sampled equidistantly, using a grid spacing of 10 pixels, to finally obtain a reasonably sized and reasonably diverse training set. Thereby, the input of each training sample consists of stacked $7 \times 7$ patches composed of (i) the backward flow components $u_{\text{bw}}$ and $v_{\text{bw}}$, (ii) a validity flag $\{0,1\}$ indicating if the location surpassed outlier filtering step and (iii) the $x$- and $y$-component of the pixel location (normalized to $[-1,1] \times [-1,1]$). The corresponding output is given by the stacked forward flow components $u_{\text{fw}}$ and $v_{\text{fw}}$. The whole process is illustrated Fig. 2 (left). Please note that the training data is extracted automatically per image triplet during the estimation and does not rely on any kind of ground truth information nor manually labeled training data.

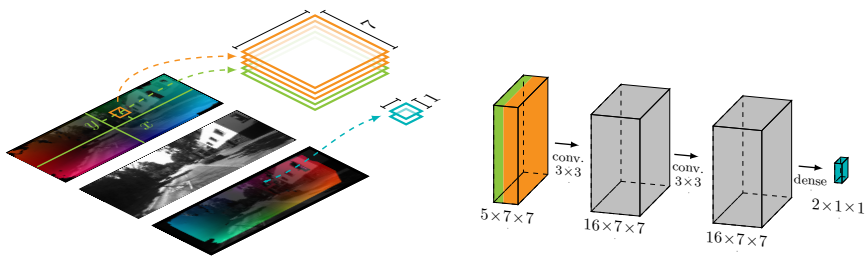

Figure 2: **Left:** Training sample extraction. **Right:** Regression network architecture.

**CNN-based Regression.** With the extracted training samples we now train a motion model in terms of a CNN which allows us to predict the forward flow solely based on the backward flow. The input of the network consists of stacked $7 \times 7$ patches including information on the backward flow, the validity and the location as described in the previous section. The output of the network is the predicted forward flow for the center location of the input patch. By considering not only the backward flows in the input patch but also the corresponding image coordinates, the network is enabled to learn a *location dependent model*. This aspect is particularly important, since motion patterns may locally vary due to independently moving objects, non-rigid deformations as well as perspective effects.

Let us now detail on the architecture and the training process of our regression network. As loss function we minimize the absolute difference of the predicted flow vector and the actual forward flow vector. Thereby, we kept the network architecture simple, since it has to be trained online for each frame of the sequence: it consists of 2 convolutional layers each with 16 kernels of window size $3 \times 3$ and a fully connected layer with a 2-vector output, which represents the desired predicted forward flow vector; see Fig. 2 (right). As non-linearities we employed ReLUs [23]. The network is implemented in TensorFlow [1] and trained using the ADAM optimizer [18] with an exponential learning rate decay. The initial learning rate is set to 0.01 and decays every 200 steps with a base of 0.8. Using the described network and learning scheme 4000 steps where sufficient to train the network.

## 2.4 Combination and Final Estimation

After learning the motion model in terms of a CNN, we can use it to predict a new forward flow based on the filtered backward flow. The predicted flow vectors can then be employed to augment the initial forward flow at those locations where no flow vectors are present; see Fig. 3. Since the combined flow field is not dense – at some locations neither forward nor backward flow vectors are available – we finally perform inpainting and refinement with the same techniques as in our baseline; i.e. RIC [13] and OIR [19].

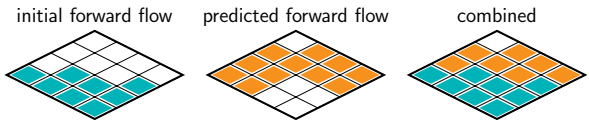

Figure 3: Illustration showing the combination step.

# 3 Experiments

To investigate the benefit of our new optical flow approach, which we named ProFlow ("**pr**edict **o**ptical **flow**"), we consider the training data sets as well as the test data sets of the three most popular optical flow benchmarks: the KITTI 2012 benchmark [11], the KITTI 2015 benchmark [22] and the MPI Sintel benchmark [6].

**Baseline Performance.** Since our baseline approach for computing the initial flow fields is not based on a single approach but on a combination of recently published techniques, we first evaluate and compare its performance with those of the original methods. The outcome is listed in Tab. 1 (top). While the results already show some improvements compared to CPM-Flow [12], RIC-Flow [13] and CPM+OIR [19], only the DF+OIR [19] approach performs slightly better. In the latter case the more advanced DiscreteFlow matches [21] are used that are, however, computationally much more expensive than our CPM-matches [12].

**Learned vs. Constant Model.** In our second experiment, we compare our learned motion model with the constant motion model that is frequently used in the literature; see e.g. [16, 17, 33]. This model assumes the forward flow $\mathbf{w}_{\text{fw}}$ and the backward flow $\mathbf{w}_{\text{bw}}$ to be simply related via $\mathbf{w}_{\text{fw}} = -\mathbf{w}_{\text{bw}}$. As already mentioned, this model can be a reasonable approximation in case of slowly moving objects [16], but it typically does not hold for fast or complex motion scenarios [32, 33]. Moreover, due to the projection involved in the optical flow, such a constant motion model does not represent an actual constant 3D motion unless the motion is parallel to the image plane. For our comparison we computed the results for the training data sets of all three benchmarks using our approach as well as a modified version, where we omitted the model learning part and directly applied the constant motion model. In Tab. 1 (bottom) we listed the outcome of both approaches. As one can see, using the constant model for predicting the optical flow does not work well for all benchmarks and even leads to a deterioration compared to the baseline. Our approach, in contrast, learns an appropriate motion model and consistently achieves improvements ranging from 8 to 27 percent. This observation is confirmed by the visual comparison in Figs. 4 and 5 that show the three input frames, the computed flow field and a bad pixel visualization for a sequence of the KITTI 2015 and the MPI Sintel benchmark, respectively. While Fig. 4 makes the quan-titative gains for the KITTI benchmark explicit, Fig. 5 shows that our approach also allows to obtain improvements in non-occluded areas, such as for the head of the dragon.

Table 1: Results for the training data sets of the KITTI 2012 benchmark [11], the KITTI 2015 benchmark [22] and the MPI Sintel benchmark [6] (clean render path) in terms of the average endpoint error (AEE) and the percentage of bad pixels (BP) with a 3px threshold.

| Method | Model | KITTI 2012 AEE | KITTI 2012 BP | KITTI 2015 AEE | KITTI 2015 BP | Sintel AEE |
|---|---|---|---|---|---|---|
| CPM-Flow [12] | – | 3.00 | 14.58 | 7.78 | 22.86 | 2.00 |
| RIC-Flow [13] | – | 2.94 | 10.94 | 7.24 | 21.46 | 2.16 |
| CPM+OIR [19] | – | 2.78 | 9.68 | 7.36 | 19.21 | 1.99 |
| DF+OIR [19] | – | **2.34** | 9.29 | **5.89** | **18.10** | **1.91** |
| our baseline | – | 2.61 | **8.98** | 6.82 | 18.70 | 1.95 |
| only prediction, no refinement | constant | 7.99 | 57.13 | 12.81 | 52.19 | 5.32 |
| only prediction | constant | 7.07 | 45.07 | 12.23 | 46.15 | 4.97 |
| only prediction, no refinement | learned | 2.27 | 7.79 | 5.87 | 17.42 | 2.93 |
| only prediction | learned | **1.83** | **7.44** | **5.37** | **16.98** | **2.29** |
| our approach | constant | 4.07 | 16.33 | 8.53 | 23.23 | 2.82 |
| our approach | learned | **1.89** | **7.26** | **5.22** | **16.25** | **1.78** |
| improvement w.r.t. baseline (%) | – | 27.5 | 19.1 | 23.4 | 13.1 | 8.7 |

**Only Prediction.** To further investigate the quality of the predicted flow fields, we performed a third experiment, where we skipped the combination step and only used the predicted flow to compute the final flow estimate. Thereby we computed the final estimate in two ways: once by solely inpainting the predicted flow field, i.e. without refinement, and once with the entire pipeline, i.e. with inpainting and refinement. The outcome is listed in Tab. 1 (middle). As one can see, in case of the KITTI 2012 and the KITTI 2015 benchmark, the pure prediction variant even outperforms our baseline. This not only confirms the high quality and reliability of our learned motion models but also reveals that due to the dominating forward motion in the benchmark many occlusions appear at the image boundaries and hence can be resolved by considering information from the preceding frame. The more challenging MPI Sintel benchmark, in contrast, does not contain such a regular motion. Nevertheless, also in this case the learned prediction is still able to achieve reasonable results. For the sake of completeness, we also computed predictions based on the constant motion model. However, as one can see, it does not allow to achieve nearly as good results as the learned approach.

**Comparison to the Literature.** In our final experiment we compare the performance of our novel optical flow approach to other methods from the literature. To this end, we submitted results for the test data sets to all three benchmarks. The results are shown in Tab. 2, where we have listed the ten best performing non-anonymous optical flow methods for each benchmark. In case of KITTI 2012 our approach ranks eighth w.r.t. the bad pixel error accounting only for pixels in *non-occluded* areas (Out-Noc). Since our method aims at improving the estimation in occluded areas, however, the bad pixel measure considering *all* pixels (Out-All) is more informative. Here, our approach ranks second. In case of the more challenging KITTI 2015 benchmark we also rank eighth. However, on the most challenging and diverse benchmark, the MPI Sintel benchmark, we rank first in the final and second in the clean render path. In particular, in the significantly more challenging final render path, we not only obtain the best result, bus also obtain the lowest error in occluded areas (unmatched) – even outperforming recent multi-frame methods such as MR-Flow [34] that combine geometric

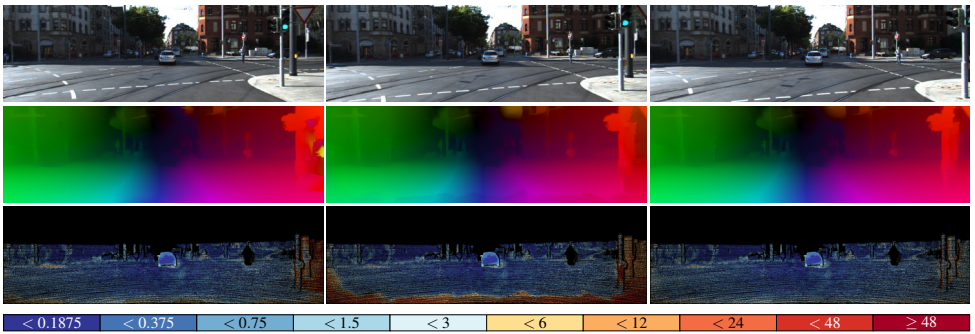

Figure 4: Example for the KITTI 2015 benchmark [22] (#149). **First row:** Previous, reference and subsequent frame. **Second and third row:** Estimated flow field, bad pixel visualization. **From left to right:** Baseline, constant motion model and our approach.

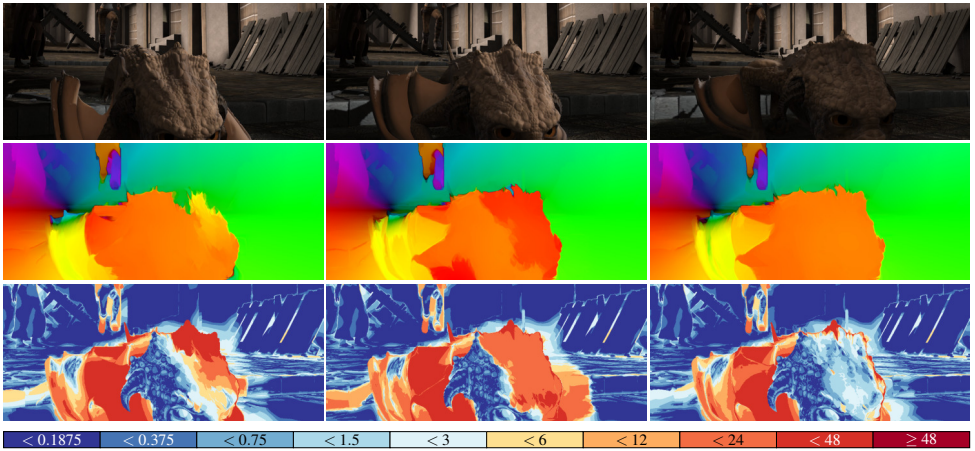

Figure 5: Example for the MPI Sintel benchmark [6] (market_5 #8). **First row:** Previous, reference and subsequent frame. **Second and third row:** Estimated flow field, bad pixel visualization. **From left to right:** Baseline, constant motion model and our approach.

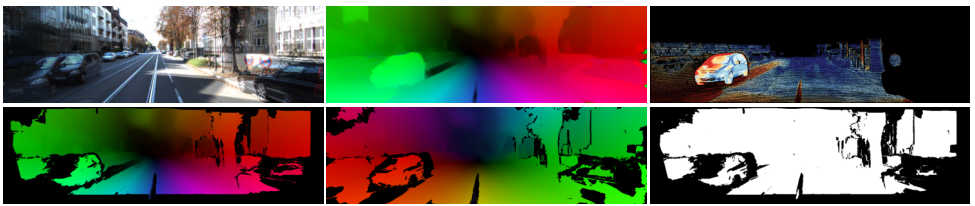

Figure 6: Limitations example (#0 KITTI 2015 benchmark [22]). **First row:** Overlayed reference and subsequent input frame, final flow estimate, bad pixel visualization. **Second row:** Filtered forward flow, filtered backward flow, possible training candidates (white).

Table 2: Top 10 non-anonymous optical flow methods on the test data of the KITTI 2012/2015 [11, 22] and of the MPI Sintel benchmark [6], excluding scene flow methods.

| KITTI 2012 | Out-Noc | Out-All | Avg-Noc | Avg-All | KITTI 2015 | Fl-bg | Fl-fg | Fl-all |
|---|---|---|---|---|---|---|---|---|
| SPS-Fl[1] | 3.38 % | 10.06 % | 0.9 px | 2.9 px | PWC-Net | 9.66 % | 9.31 % | 9.60 % |
| PCBP-Flow[1] | 3.64 % | 8.28 % | 0.9 px | 2.2 px | MirrorFlow | 8.93 % | 17.07 % | 10.29 % |
| SDF[2] | 3.80 % | 7.69 % | 1.0 px | 2.3 px | SDF[2] | 8.61 % | 23.01 % | 11.01 % |
| MotionSLIC[1] | 3.91 % | 10.56 % | 0.9 px | 2.7 px | UnFlow | 10.15 % | 15.93 % | 11.11 % |
| PWC-Net | 4.22 % | 8.10 % | 0.9 px | 1.7 px | CNNF+PMBP | 10.08 % | 18.56 % | 11.49 % |
| UnFlow | 4.28 % | 8.42 % | 0.9 px | 1.7 px | MR-Flow[2] | 10.13 % | 22.51 % | 12.19 % |
| MirrorFlow | 4.38 % | 8.20 % | 1.2 px | 2.6 px | DCFlow | 13.10 % | 23.70 % | 14.86 % |
| **our approach** | **4.49 %** | **7.88 %** | **1.1 px** | **2.1 px** | **our approach** | **13.86 %** | **20.91 %** | **15.04 %** |
| ImpPB+SPCI | 4.65 % | 13.47 % | 1.1 px | 2.9 px | SOF[2] | 14.63 % | 22.83 % | 15.99 % |
| CNNF+PMBP | 4.70 % | 14.87 % | 1.1 px | 3.3 px | JFS[2] | 15.90 % | 19.31 % | 16.47 % |
| **MPI Sintel final** | **all** | **matched** | **unmatched** | | **MPI Sintel clean** | **all** | **matched** | **unmatched** |
| **our approach** | **5.017** | **2.596** | **24.736** | | MR-Flow[2] | 2.527 | 0.954 | 15.365 |
| PWC-Net | 5.042 | 2.445 | 26.221 | | **our approach** | **2.818** | **1.027** | **17.428** |
| DCFlow | 5.119 | 2.283 | 28.228 | | FlowFields+ | 3.102 | 0.820 | 21.718 |
| FlowFieldsCNN | 5.363 | 2.303 | 30.313 | | CPM2 | 3.253 | 0.980 | 21.812 |
| MR-Flow[2] | 5.376 | 2.818 | 26.235 | | MirrorFlow | 3.316 | 1.338 | 19.470 |
| S2F-IF | 5.417 | 2.549 | 28.795 | | DF+OIR | 3.331 | 0.942 | 22.817 |
| InterpoNet_ff | 5.535 | 2.372 | 31.296 | | S2F-IF | 3.500 | 0.988 | 23.986 |
| RicFlow | 5.620 | 2.765 | 28.907 | | SPM-BPv2 | 3.515 | 1.020 | 23.865 |
| InterpoNet_cpm | 5.627 | 2.594 | 30.344 | | DCFlow | 3.537 | 1.103 | 23.394 |
| ProbFlowFields | 5.696 | 2.545 | 31.371 | | RicFlow | 3.550 | 1.264 | 22.220 |

[1] uses epipolar geometry as a hard constraint, only applicable to pure ego-motion
[2] exploits semantic information

constraints with a semantic rigidity segmentation. This shows that, in particular in difficult scenes with partially non-rigid motion, learned temporal models may be worthwhile strategy.

**Runtime and Parameters.** Running our approach on a desktop PC equipped with an Intel Core i7-7820X CPU @ 3.60GHz and an Nvidia GeForce GTX 1070 the runtime is approximately 112s for a flow field of size 1226×370. The overall runtime splits up into: 36s for the initial flow field estimation, 50s for the motion model learning and prediction, and another 26s for the final inpainting and refinement. Regarding the parameters of the used approaches (CPM, RIC, OIR), we used the default parameters as provided by the authors [12, 13, 19]. In case of the matching (CPM) and inpainting (RIC) these parameters are the same for all benchmarks, only in case of the refinement (OIR) there is a set of parameters per benchmark.

**Limitations.** Finally, we also want to comment on the limitations of our approach. In case of large image regions that only contain poor or possibly no training samples, the validity of the learned motion model may not be able to generalize to the entire image domain. In Fig. 6 such a scenario is depicted. Due to the missing training samples at the bottom corners of the image, the prediction cannot achieve a noticeable improvement in these areas. This problem, however, could be resolved by additionally using geometric constraints in terms of a rigid motion model. Hence, we believe that combining our learning based approach with such a model could even allow for further improvements – at least in case of rigid scenes with a vast amount of ego-motion, such as the KITTI 2012 and the KITTI 2015 benchmark.

## 4 Conclusions

We have presented a novel multi-frame optical flow approach that integrates flow predictions based on a CNN. To this end, we made use of an unsupervised learning approach that learns a motion model by estimating a spatially variant mapping from the backward to the forward flow. In contrast to existing approaches from the literature that train their network only once before the estimation based on a huge data set, our methods exploits flow estimates from the current image sequence to learn the model online, i.e. during the estimation. In this way, it becomes possible to learn motion models that are specifically tailored to the actual motion occurring in each frame. Experiments made the good performance of our method explicit. They not only show significant improvements compared to a baseline without prediction, they also show consistently good results in all major benchmarks – including top results on the MPI Sintel benchmark.

**Acknowledgments.** We thank the German Research Foundation (DFG) for financial support within project B04 of SFB/Transregio 161. Moreover, we thank Lourdes Agapito for helpful discussions.

12 MAURER, BRUHN: PROFLOW: LEARNING TO PREDICT OPTICAL FLOW[24] A. Ranjan and M. J. Black. Optical flow using a spatial pyramid network. In *Proc. IEEE Conference on Computer Vision and Pattern Recognition*, pages 2720–2729, 2017.

[25] Z. Ren, J. Yan, B. Ni, B. Liu, X. Yang, and H. Zha. Unsupervised deep learning for optical flow estimation. In *Proc. AAAI Conference on Artificial Intelligence*, 2017.

[26] J. Revaud, P. Weinzaepfel, Z. Harchaoui, and C. Schmid. Epicflow: Edge-preserving interpolation of correspondences for optical flow. In *Proc. IEEE Conference on Computer Vision and Pattern Recognition*, pages 1164–1172, 2015.

[27] S. Ricco and C. Thomasi. Dense Lagrangian motion estimation with occlusions. In *Proc. IEEE Conference on Computer Vision and Pattern Recognition*, pages 1800–1807, 2012.

[28] H. Sawhney. 3D geometry from planar parallax. In *Proc. IEEE Conference on Computer Vision and Pattern Recognition*, pages 929–934, 1994.

[29] L. Sevilla-Lara, D. Sun, V. Jampani, and M. J. Black. Optical flow with semantic segmentation and localized layers. In *Proc. IEEE Conference on Computer Vision and Pattern Recognition*, pages 3889–3898, 2016.

[30] D. Sun, S. Roth, J. P. Lewis, and M. J. Black. Learning optical flow. In *Proc. European Conference on Computer Vision*, pages 83–97, 2009.

[31] D. Sun, X. Yang, M. Y. Liu, and J. Kautz. PWC-Net: CNNs for optical flow using pyramid, warping, and cost volume. In *Proc. IEEE Conference on Computer Vision and Pattern Recognition*, 2018. To appear.

[32] S. Volz, A. Bruhn, L. Valgaerts, and H. Zimmer. Modeling temporal coherence for optical flow. In *Proc. International Conference on Computer Vision*, pages 1116–1123, 2011.

[33] M. Werlberger, W. Trobin, T. Pock, A. Wedel, D. Cremers, and H. Bischof. Anistropic Huber-L1 optical flow. In *Proc. British Machine Vision Conference*, pages 1–11, 2009.

[34] J. Wulff, L. Sevilla-Lara, and M. J. Black. Optical flow in mostly rigid scenes. In *Proc. IEEE Conference on Computer Vision and Pattern Recognition*, pages 6911–6920, 2017.

[35] J. Xu, R. Ranftl, and V. Koltun. Accurate optical flow via direct cost volume processing. In *Proc. IEEE Conference on Computer Vision and Pattern Recognition*, pages 5807–5815, 2017.

[36] J. J. Yu, A. W. Harley, and K. G. Derpanis. Back to basics: Unsupervised learning of optical flow via brightness constancy and motion smoothness. In *Proc. Workshops European Conference on Computer Vision*, pages 3–10, 2016.

[37] Y. Zhu, Z. Land, S. Newsam, and A. G. Hauptmann. Guided optical flow learning. In *Proc. IEEE Workshops Conference on Computer Vision and Pattern Recognition*, 2017.

# A  Visual Results

To make the improvements of our approach more graspable, we depict a selection of sequences that contain a variety of different scenarios, i.a. such as non-rigid deformations, occlusions and out-of-frame motion. These can be found in Fig. 7 – 9 (KITTI 2015) and Fig. 10 – 13 (MPI Sintel). On the one hand, one can see from visualization of selected areas that predictions by the learned motion models (orange) are mainly used in occluded regions while the initial forward flow (turquois) is used elsewhere. On the other hand, one can observe from the baseline and the final flow field (our approach) that this typically leads to a noticeable improvement of the estimation quality.

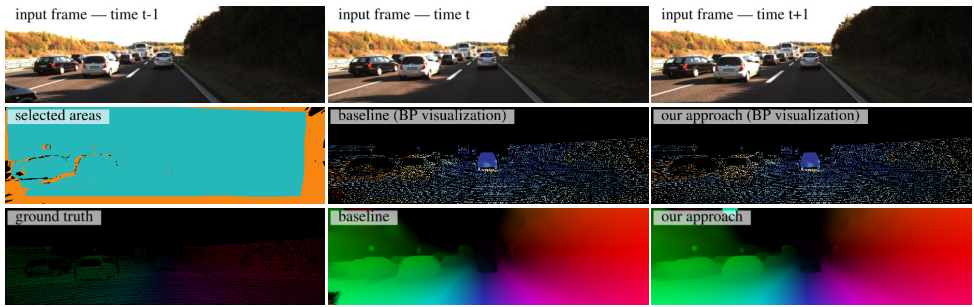

Figure 7: Visualization of improvements for Sequence #199 of KITTI 2015 [22].

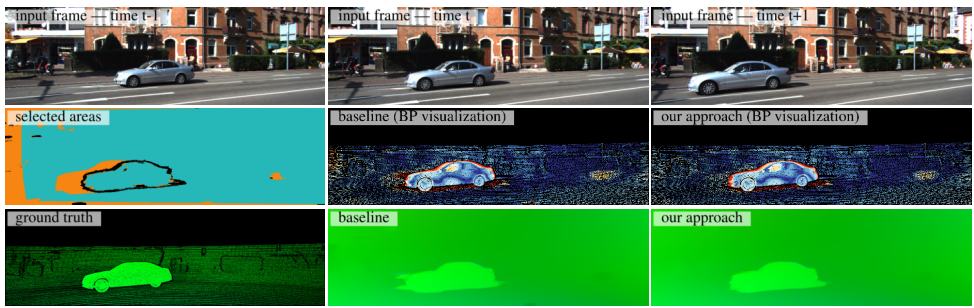

Figure 8: Visualization of improvements for Sequence #133 of KITTI 2015 [22].

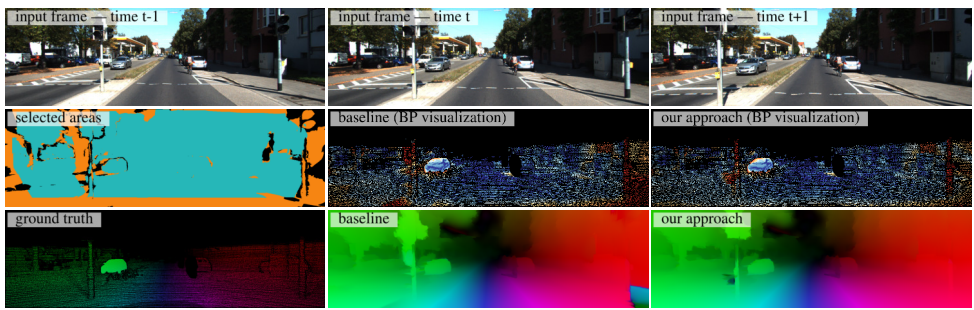

Figure 9: Visualization of improvements for Sequence #186 of KITTI 2015 [22].

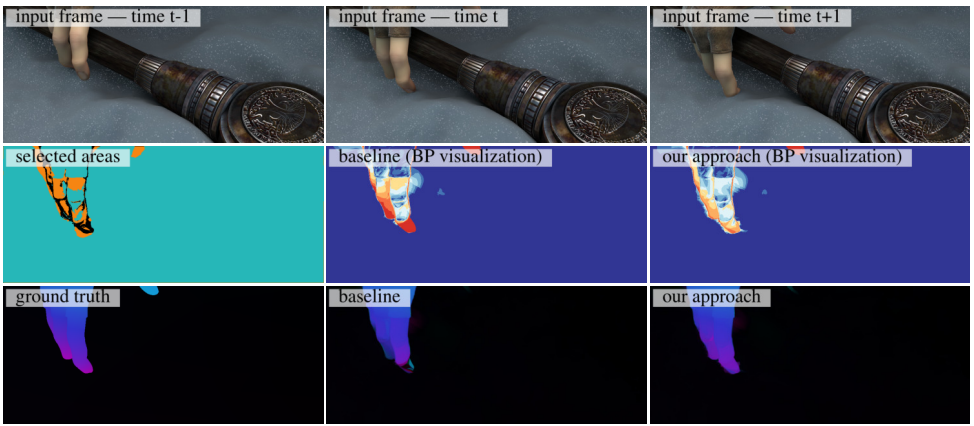

Figure 10: Visualization of improvements for the ambush_7 sequence (#9) of MPI Sintel [6].

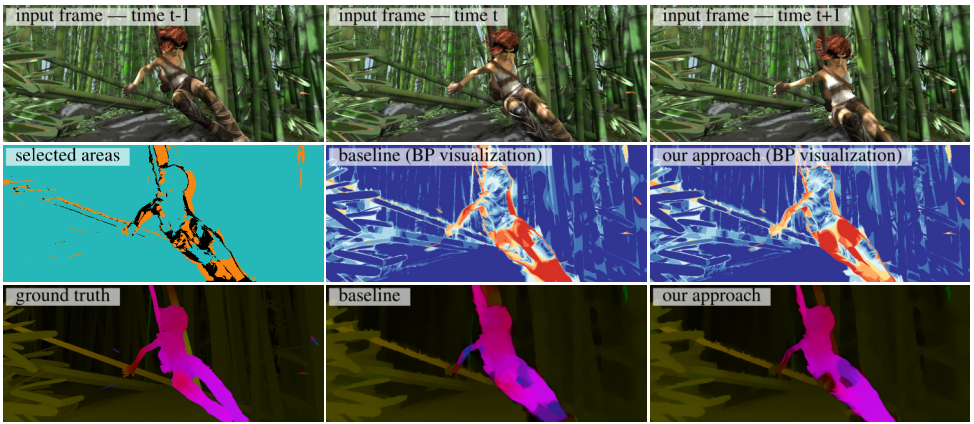

Figure 11: Visualization of improvements for the bamboo_2 sequence (#43) of MPI Sintel [6].

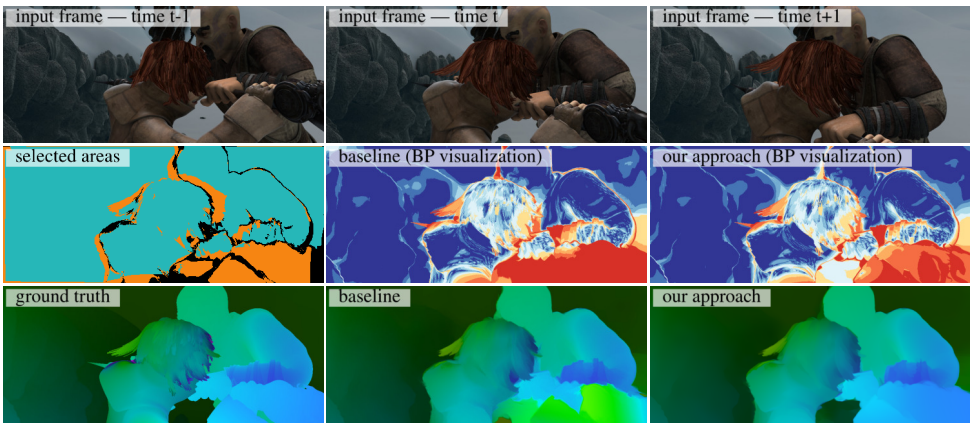

Figure 12: Visualization of improvements for the ambush_5 sequence (#31) of MPI Sintel [6].

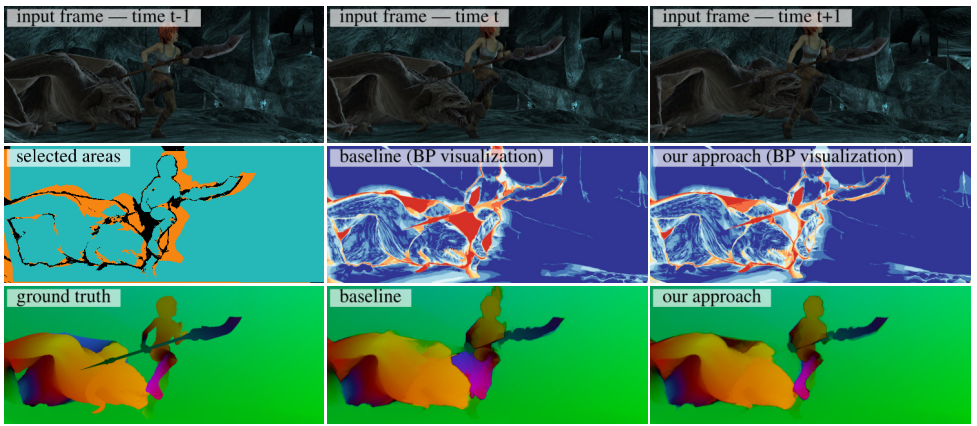

Figure 13: Visualization of improvements for the cave_2 sequence (#13) of MPI Sintel [6].

**(i) Input Frames**

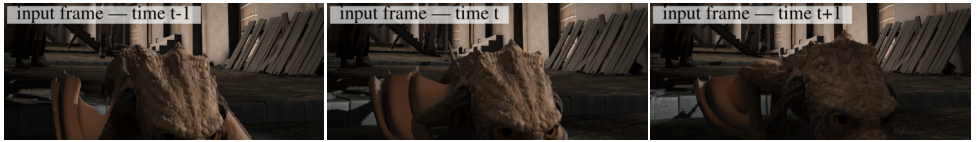

**(ii) Initial Flow Estimation & Outlier Filtering**

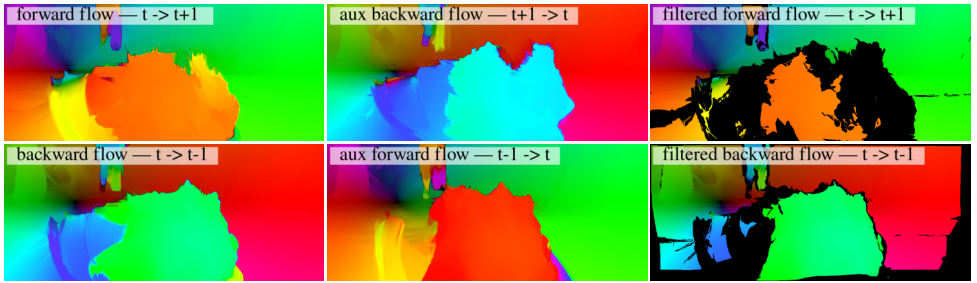

**(iii) Model Learning & Prediction**

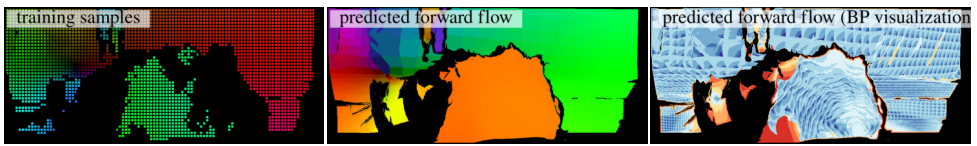

**(iv) Combination & Final Estimation**

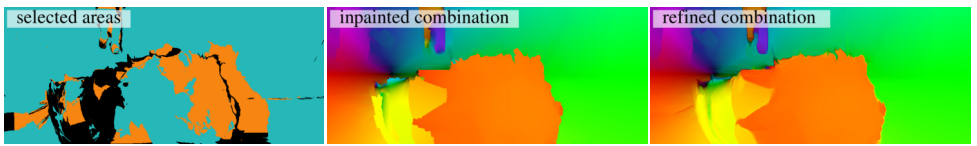

Figure 14: Overview showing intermediate results of our optical flow approach for the market_5 #8 sequence of the MPI Sintel benchmark [6].

(i) Input Frames

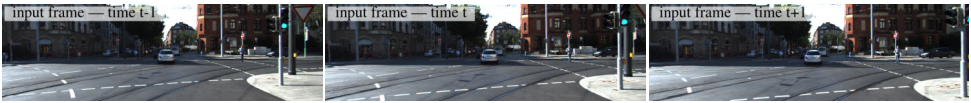

(ii) Initial Flow Estimation & Outlier Filtering

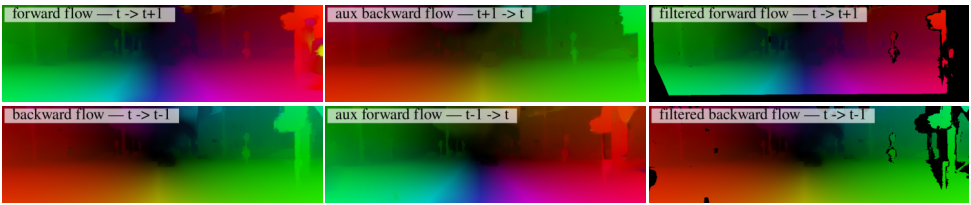

(iii) Model Learning & Prediction

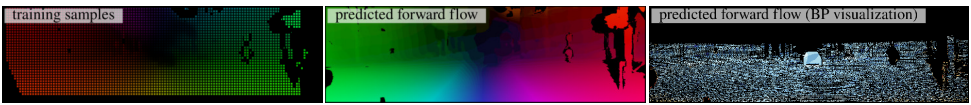

(iv) Combination & Final Estimation

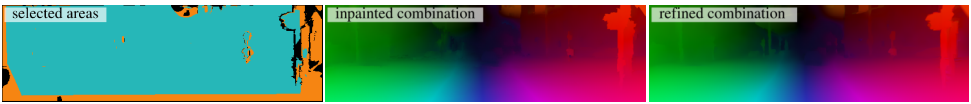

Figure 15: Overview showing intermediate results of our optical flow approach for the #149 sequence of the KITTI 2015 benchmark [22].

# B  Visualization of Intermediate Results

In order to provide a better insight into the overall approach, we show intermediate results of the different steps for the two exemplary sequences depicted in the main paper. These intermediate results are presented in Fig. 14 and Fig. 15. They include:

(i) the three input images with the reference frame at time *t*

(ii) the initial forward and backward flow fields, the auxiliary flow fields (required to perform the bi-directional consistency check), as well as the filtered flow field

(iii) the considered training samples, the predicted forward flow (computed using the backward flow and the learned motion model), and a bad pixel (BP) visualization of the predicted flow field

(iv) the decisions within the combination step (turquois=forward, orange=prediction), the inpainted combined flow field and the final flow estimate (refined combination)